\documentclass{article}
\usepackage{arxiv}
\usepackage{cite}

\usepackage{graphicx}
\usepackage[dvipsnames]{xcolor} 
\usepackage{pifont} 

\usepackage{graphicx}
\usepackage{multirow}
\usepackage{amsmath,amssymb,amsfonts,amsthm}
\usepackage{bm}
\usepackage{mathrsfs}
\usepackage{mathtools}
\usepackage{bbm}
\usepackage[title]{appendix}
\usepackage[dvipsnames]{xcolor}
\usepackage{textcomp}
\usepackage{manyfoot}
\usepackage{booktabs}
\usepackage{url}
\usepackage[algo2e,ruled,vlined,linesnumbered]{algorithm2e}
\usepackage{listings}
\usepackage{adjustbox}
\usepackage{cases}
\usepackage{bbding}
\usepackage{pifont}
\usepackage{circuitikz}
\usepackage{etoolbox} %
\usepackage{booktabs,tabularx} 
\usepackage{dsfont}
\usepackage{enumitem}

\DeclareMathOperator{\sgn}{sign}

\newcommand{\R}{\mathds{R}}
\newcommand{\Z}{\mathds{Z}}

\theoremstyle{thmstylethree}
\newtheorem{definition}{Definition}

\newcommand{\set}[1]{\left\{#1\right\}}
\newcommand{\br}[1]{\left(#1\right)}

\renewcommand{\epsilon}{\varepsilon}

\renewcommand{\rm}[1]{\ensuremath{\mathrm{#1}}}

\newcommand{\unum}[1]{{^{\br{#1}}}}

\let\oldnl\nl
\newcommand{\nlnonumber}{\renewcommand{\nl}{\let\nl\oldnl}}

\newcommand{\uone}[1]{{#1}}

\AtBeginDocument{%
  \providecommand\BibTeX{{%
    \normalfont B\kern-0.5em{\scshape i\kern-0.25em b}\kern-0.8em\TeX}}}

\begin{document}

\title{Lens-descriptor guided evolutionary algorithm for optimization of complex optical systems with \\ glass choice}
\author{\begin{tabular}{c c c}
    Kirill Antonov$^{1}$ & Teus Tukker$^{2}$ & Tiago Botari$^{2}$  \\
    Thomas H.~W. Bäck$^{1}$  & Anna V. Kononova$^{1}$  & Niki van Stein$^{1}$
\end{tabular}\\[5mm]
$^{1}$Leiden University, The Netherlands\\
$^{2}$ASML, Veldhoven, The Netherlands\\
}

\maketitle

\begin{abstract}
Designing high-performance optical lenses entails exploring a high-dimensional, tightly constrained space of surface curvatures, glass choices, element thicknesses, and spacings. In practice, standard optimizers (e.g., gradient-based local search and evolutionary strategies) often converge to a single local optimum, overlooking many comparably good alternatives that matter for downstream engineering decisions. We propose the \emph{Lens Descriptor-Guided Evolutionary Algorithm} (LDG-EA), a two-stage framework for multimodal lens optimization. LDG-EA first partitions the design space into \emph{behavior descriptors} defined by curvature-sign patterns and material indices, then learns a probabilistic model over descriptors to allocate evaluations toward promising regions. Within each descriptor, LDG-EA applies the Hill-Valley Evolutionary Algorithm with covariance-matrix self-adaptation to recover multiple distinct local minima, optionally followed by gradient-based refinement. On a 24-variable (18 continuous and 6 integer), six-element Double-Gauss topology, LDG-EA generates on average \(14.5\times 10^{3}\) candidate minima spanning 636 unique descriptors, an order of magnitude more than a CMA-ES baseline, while keeping wall-clock time at one hour scale. Although the best LDG-EA design is slightly worse than a fine-tuned reference lens, it remains in the same performance range. Overall, the proposed LDG-EA produces a diverse set of solutions while maintaining competitive quality within practical computational budgets and wall-clock time.
\end{abstract}


\section{Introduction}

\begin{figure*}[!tb]
  \centering
  \begin{tabular}{ p{0.45\textwidth} p{0.5\textwidth}}
    \includegraphics[width=1\linewidth, trim=0mm 40mm 00mm 00mm]{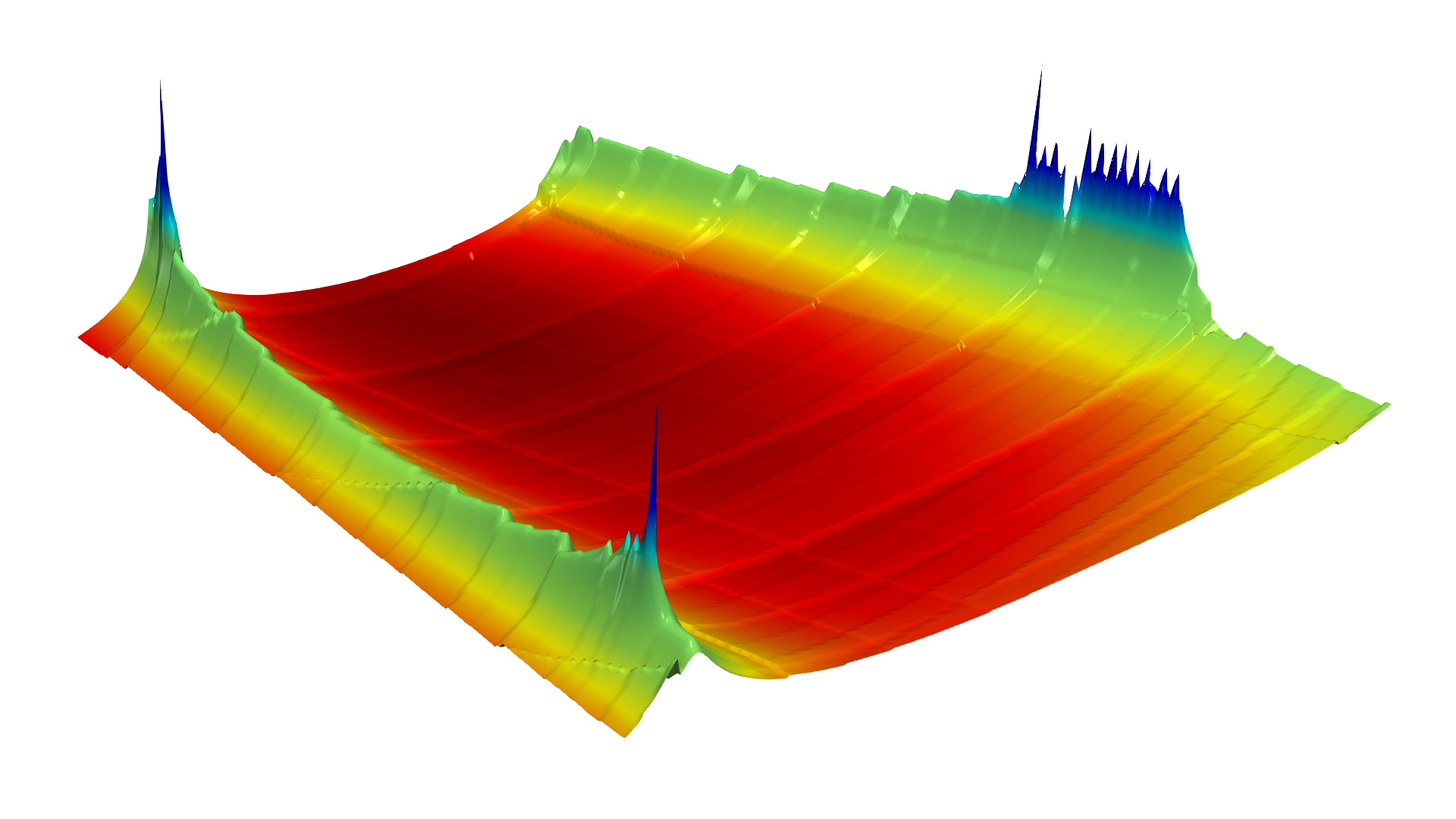} &
    \includegraphics[width=1\linewidth, trim=0mm 40mm 00mm 00mm]{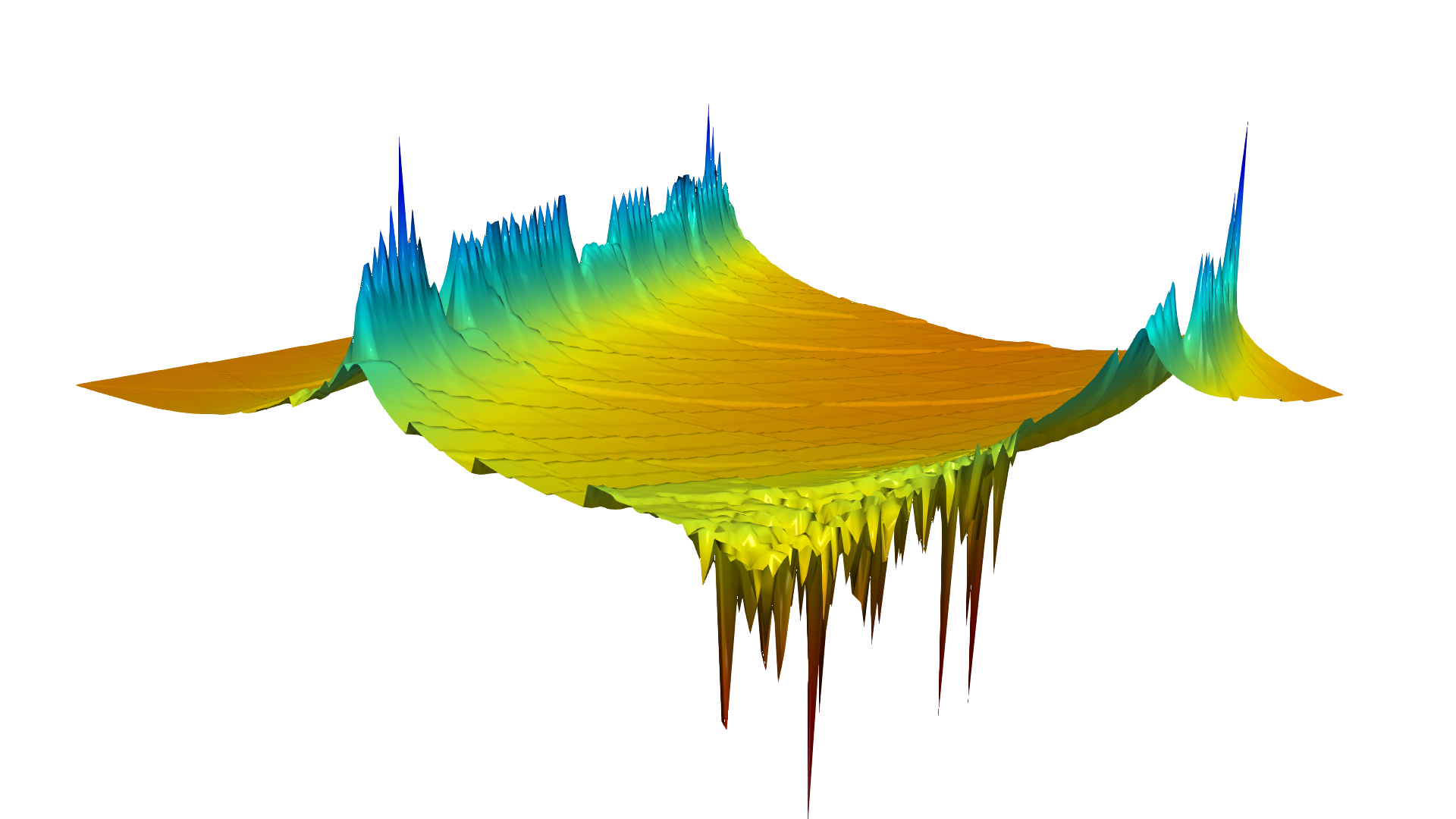}
    \\
  \end{tabular}
  \caption{Two views of the optimization landscape for a four‐element camera lens, where variables are curvatures, distances, and materials of glasses, \uone{visualized using the method from work~\cite{antonov2024quality}}. \uone{This method selects points between three local minima, identified through a local-search algorithm.} The horizontal axes correspond to a plane in the high-dimensional design space, and the vertical axis shows the reversed merit function (the higher, the better). \uone{Multiple optima indicate distinct families of high-quality designs.}}
  \label{fig:landscape}
\end{figure*}

Optical lens systems enable a broad spectrum of advanced technologies, from smartphone and machine‐vision cameras to the ultra‐precise optics used in 
lithography and metrology equipment~\cite{asmlreport2022}. 
Designing these and similar systems involves selecting optical surface curvatures, glass types, element thicknesses, and inter‐element spacings to meet key performance metrics, such as root mean square (RMS) spot size, focal length, and field number while respecting manufacturing constraints on element size, weight, and material availability~\cite{o1985elements,laikin2018lens,yabe2018optimization}. 
The resulting design space is both high‐dimensional and highly multimodal~\cite{das2011real}, with numerous local optima in the objective function~\cite{turnhout2009instabilities}. 
Figure~\ref{fig:landscape} illustrates this phenomenon on a two-dimensional slice through a real lens-design landscape: multiple separated peaks represent different high-performing \emph{families} of designs rather than minor perturbations of a single solution.

In industrial practice, optical engineers typically work in a designer-in-the-loop manner: they propose a starting layout (e.g., from patents or their own expertise) and then apply local numerical optimization using commercial software (e.g., OpticStudio~\cite{opticsstudio} or CODE V~\cite{CODEV2021}).
This workflow is highly effective for refinement, but it is biased toward only several attraction basins and therefore explores only a tiny fraction of the design space.
CODE V's Global Synthesis (GS) module~\cite{globalsynth} is a widely used alternative intended to broaden exploration.
However, GS provides limited control over the diversity of the returned solutions and offers only partial support for systematic glass optimization, making it difficult to deliberately generate a varied portfolio of high-quality designs.

The ability to produce \emph{multiple, diverse} high-quality designs is practically valuable.
A diverse solution set supports downstream engineering choices that cannot be reliably incorporated into the optimization itself, including validation of previously unknown structural constraints, consideration of glass availability, and selection of designs with advantageous cost, manufacturability, and tolerance characteristics.
Motivated by these needs, a broad range of learning-based methods~\cite{yow2024artificial} and metaheuristics~\cite{thibault2005evolutionary,hoschel2019genetic} have been proposed.
Neural networks have been used to extrapolate reference designs taken from the literature~\cite{cote2019extrapolating,cote2021deep,cote2022inferring}, and surrogate models combined with Differential Evolution~\cite{das2010differential} can reduce expensive ray-trace evaluations (e.g., by around 10\%)~\cite{hegde2019accelerating}.
Deep models have also been explored for tolerancing and uncertainty quantification~\cite{li2022novel,shahane2022surrogate} and for modeling specialized aberration behavior~\cite{zhang2023uniform}.
From the evolutionary computation perspective, classical global optimization methods have been successfully applied to lens design~\cite{nagata2004lens,menke2018application,qian2023evolved}.
Despite these advances, these methods are still limited in glass selection and exploration since they do not exploit the specific structure of lens-design landscapes and therefore do not aim to systematically cover diverse high-quality designs.

Multimodal optimization methods offer a natural framework for addressing this problem, as they are explicitly designed to identify multiple optima within a single run~\cite{preuss2021metaheuristics}.
Recent studies have indeed investigated multimodal optimization for optical design~\cite{kononova2021addressing,kononova2021locating,antonov2023new,antonov2025quality}.
However, the existing applications of multimodal methods optimize only surface curvatures, which limits practical relevance because glass choice is a crucial element of real designs.
Moreover, reported wall-clock times span multiple days to weeks even on moderate lens systems (three to four elements), whereas industrial tools such as GS can often deliver solutions within about an hour on such systems on the same machine~\cite{antonov2025quality}.

In this paper, we close this gap by introducing a multimodal evolutionary algorithm that optimizes glasses, curvatures, element thicknesses, and inter-element spacings for a practical six-element Double-Gauss lens system under an hour-scale budget.
We propose the \emph{Lens Descriptor-Guided Evolutionary Algorithm} (LDG-EA), a two-stage framework that explicitly exploits domain structure by partitioning designs into interpretable \emph{behavior descriptors} and learning an adaptive sampling distribution over them.
At each iteration, LDG-EA samples promising descriptors, applies the Hill-Valley Evolutionary Algorithm~\cite{maree2018real} to locate local optima within the subspace induced by each descriptor, and updates descriptor probabilities based on the observed performance, balancing exploration of new regions with exploitation of high-quality design regions.

Several leading multimodal optimizers also allocate search effort across landscape regions, for example BIPOP-CMA-ES~\cite{hansen2009benchmarking}, DPI-CMA-ES~\cite{Shir-NACO08}, DPI-MIES~\cite{li2008mixed}, LADE~\cite{lin2025landscape} and NetCDE~\cite{chen2023network}.
In contrast to these largely domain-agnostic partitions, LDG-EA defines \emph{lens-specific} niches a priori through behavior descriptors that capture curvature and glass patterns.
This descriptor-driven partition \emph{explicitly} enforces diversity across design patterns. 
As a result, LDG-EA balances between solution quality and diversity to enable robust multimodal search tailored to lens design.

\paragraph{Contributions.}
\begin{itemize}
 \item We formalize lens-design behavior descriptors and use them to structure and navigate the lens-design multimodal optimization landscape.
 \item We develop LDG‐EA, which integrates descriptor learning, Hill-Valley Evolutionary Algorithm, and optional gradient‐based refinement into an adaptive search strategy.
 \item We validate LDG‐EA on a six‐element Double‐Gauss lens, demonstrating an order‐of‐magnitude increase in distinct high‐quality designs over a baseline algorithm and competitive RMS performance under equal computational budgets.
\end{itemize}

\section{Formalization of the optimization problem} 
\label{sec:formalization}

Let $\Theta_0\subseteq\R^{n_c}\times\R^{n_d}\times\Z^{n_m}$ be the full design space for an optical lens, and let $\mathcal{H}\subset\Theta_0$ denote a \emph{forbidden region} of restricted parameter combinations. \uone{The possible values of curvatures and thicknesses are usually restricted by manufacturing constraints.}
We define the \emph{feasible domain}
$$
\Theta \;=\;\Theta_0 \setminus \mathcal{H}.
$$
On $\Theta$, the objective function
$$
F:\Theta\;\to\;\R
$$
assigns to each design $\bm{\theta}=(\bm{\kappa},\bm{d},\bm{m})$ its optical-quality metric.

\begin{figure}[!tb]
    \centering
    \includegraphics[scale=1]{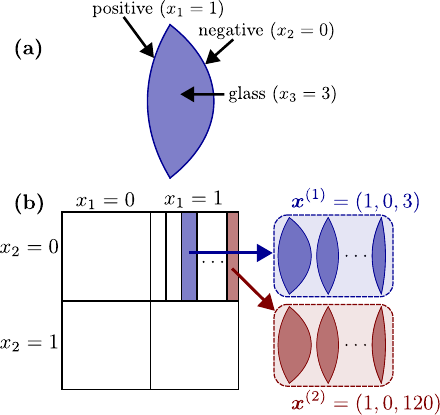}
    \caption{(a) Visualization of behavior descriptor mapping: singlet lens $\bm{\theta}$ shown in the figure is mapped to its behavior descriptor $\bm{x} = (1, 0, 3)$ by the defined mapping $\bm{x} = \mathcal{D}(\bm{\theta}).$ The glass material is represented by the color. (b) Visualization of the space of all behavior descriptors $\mathcal{X}.$ Two boxes on the right show examples of lenses with the same behavior descriptors $\bm{x}\unum{1}$ and $\bm{x}\unum{2}$ respectively. Note that singlet lenses in the same box have different thicknesses and different curvature values, but still have the same behavior descriptor due to the definition of our mapping $\mathcal{D}.$}
    \label{fig:cube}
\end{figure}

We introduce a \emph{descriptor mapping}
$$
\mathcal{D}:\Theta\;\longrightarrow\;\mathcal{X},
\qquad
\bm{\theta}\mapsto\bm{x},
$$
which retains only the qualitative features of a design. 
Namely, the signs of its surface curvatures and its discrete material indices.
Note that distance is not part of the behavior descriptor.
\uone{Each element of set $\mathcal{X}$ is called a \emph{behavior descriptors}, so the set $\mathcal{X}$ is the collection of all \emph{behavior descriptors}.}
An example of the defined behavior descriptors for singlet lenses is shown in Figure~\ref{fig:cube}.

\begin{definition}[Lens equivalence with \uone{respect} to the descriptor and performance]
\label{def:descr-eq}
Two designs $\bm{\theta}^{(1)},\bm{\theta}^{(2)}\in\Theta$ are said to be \emph{descriptor‐equivalent}, written 
$
\bm{\theta}^{(1)} \equiv \bm{\theta}^{(2)},
$
if
$$
\mathcal{D}\bigl(\bm{\theta}^{(1)}\bigr)
=\mathcal{D}\bigl(\bm{\theta}^{(2)}\bigr)
\quad\text{and}\quad
F\bigl(\bm{\theta}^{(1)}\bigr)
=F\bigl(\bm{\theta}^{(2)}\bigr).
$$
\end{definition}

In this case, $\mathcal{D}(\bm{\theta})$ is called the \emph{behavior descriptor} of $\bm{\theta}$.
Definition~\ref{def:descr-eq} \uone{follows the method for optical systems comparison from the book~\cite{braat2019imaging}. Note that, following this method, the thicknesses are not considered when comparing the similarities of optical systems.}

Our goal is to identify a set of $k$ \emph{distinct} (non-descriptor-equivalent) local optima in $\Theta$, each achieving \uone{objective value} below a threshold $c$ and each having a unique behavior descriptor.
\uone{We define the objective function in such a way that it penalizes imperfections in diffraction-limited imaging. The optimal value (although it is not always achievable in practice) of our objective function is 0, and the larger the value, the worse the performance of the system.}
Formally, find
$$
\bigl\{\bm{\theta}^{(1)},\,\bm{\theta}^{(2)},\,\dots,\,\bm{\theta}^{(k)}\bigr\}
\;\subseteq\;\Theta
$$
such that for every $i=1,\dots,k$:
\begin{enumerate}
 \item $\bm{\theta}^{(i)}$ is a local minimizer of $F$:
 $$
  \nabla F\bigl(\bm{\theta}^{(i)}\bigr)=\mathbf{0},
  \quad
  \nabla^2F\bigl(\bm{\theta}^{(i)}\bigr)\succeq 0;
 $$
 \item $F\bigl(\bm{\theta}^{(i)}\bigr)<c$;
 \item the descriptors are pairwise different:
 $$
  \mathcal{D}\bigl(\bm{\theta}^{(i)}\bigr)
  \neq
  \mathcal{D}\bigl(\bm{\theta}^{(j)}\bigr)
  \quad\forall\,i<j.
 $$
\end{enumerate}

\uone{Note that we tacitly assume that the objective function $F$ is twice continuously differentiable. Although we do not explicitly use this assumption in the following paper, we maintain this assumption because employed Evolution Strategies rely on it. We will assume that the gradient of the function $F$ is available, for example, using automatic differentiation~\cite{wang2022differentiable}.}

\section{Methodology}
\label{sec:methodology}

We introduce the \textbf{Lens Descriptor-Guided Evolutionary Algorithm (LDG-EA)}, an iterative two-stage optimization framework that dynamically focuses search effort on regions of the lens descriptor space most likely to yield high-quality designs. 

\subsection{Notation and Overview}
\label{sec:stages}
Let $\mathcal{X}$ denote the finite set of behavior descriptors, let $p\unum{t}(\bm{x})$ be a probability mass function over $\mathcal{X}$ at iteration $t$, and let $B$ be a fixed computational budget for \uone{evaluations of $F$ inside a given behavior descriptor. 
We say that the designs $\bm{x}\unum{1}, \bm{x}\unum{2}$ are evaluated inside the same behaviour descriptor if $\mathcal{D}\bigl(\bm{x}\unum{1}\bigr) = \mathcal{D}\bigl(\bm{x}\unum{2}\bigr).$}
\uone{Our algorithm uses parameters} $\lambda$ and $\mu$ to denote the number of samples and the selection size \uone{(defined in Stage 2 below)}, respectively, with $1 \le \mu \le \lambda \ll |\mathcal{X}|$. 

At each iteration $t = 1,2,3,\ldots$, the algorithm performs the following two stages:

\begin{enumerate}[leftmargin=\dimexpr0.6em+\parindent\relax]
 \item \textbf{Stage 1 -- Descriptor Sampling and Evaluation.} 
  \begin{enumerate}[leftmargin=1em]
   \item \emph{Sampling:} draw $\lambda$ descriptors
  \uone{$$
    \bm{x}\unum{t, 1}, \dots, \bm{x}\unum{t, \lambda} \;\sim\; p\unum{t}(\bm{x}).
   $$}
   \item \emph{Per-Descriptor Search Instances:} for each sampled descriptor $\bm{x}\unum{t, i}$, instantiate an \emph{internal optimization algorithm} constrained to generate solutions only \uone{inside} the descriptor $\bm{x}\unum{t, i}$ \uone{within allocated} budget $B$. 
   We denote by 
   $$
    \bm{A}\unum{t,i} \;=\;\bigl\{\bm{\theta}\unum{t,i,1},\dots,\bm{\theta}\unum{t,i,n_i}\bigr\}
   $$
   the resulting archive of \uone{all found} candidate solutions, that \uone{\emph{approximate} \emph{local minima}} at the \uone{subspace $$\mathcal{D}^{-1}\br{\bm{x}\unum{t, i}} \subset \Theta,$$ meaning such subspace that $$\forall \bm{\theta} \in \mathcal{D}^{-1}\br{\bm{x}\unum{t, i}} \implies \mathcal{D}\br{\bm{\theta}} = \bm{x}\unum{t, i}. $$}
   \item \emph{Performance Feedback:} compute the best objective value \uone{per descriptor}
   $$
    f\unum{t,i} \;=\;\min_{\,\bm{\theta}\in \bm{A}\unum{t,i}}F(\bm{\theta}),
   $$
   and record the pair $\br{\bm{x}\unum{t,i}, f\unum{t,i}}$ for descriptor‐level feedback.
  \end{enumerate}

 \item \textbf{Stage 2 -- Descriptor Selection and Distribution Update.} 
  \begin{enumerate}[leftmargin=1em]
   \item \emph{Selection:} sort the evaluated descriptors by ascending $f\unum{t,i}$ and select the top $\mu$:
   $$
    \bigl\{\bm{x}\unum{1},\dots,\bm{x}\unum{\mu}\bigr\}
    \;=\;
    \underset{i \in [1,\lambda]}{\arg\min}^{\mu}\;f\unum{t,i}.
   $$
   \item \emph{Update Rule:} update the sampling distribution $p\unum{t}(\bm{x})\to p\unum{t+1}(\bm{x})$ by increasing probability mass on the selected descriptors and decreasing it elsewhere.
   We use the following update rule, as proposed in the classical Univariate Marginal Distribution Algorithm (UMDA)~\cite{muhlenbein1996recombination}
   $$
    p\unum{t+1}(\bm{x})
    \;=\;
    (1-\alpha)\,p\unum{t}(\bm{x})
    \;+\;
    \alpha\, \prod_{i=1}^{n_c+n_m} \br{ \frac{1}{\mu}\sum_{j=1}^\mu \mathbf{1}\set{x_i=x_i\unum{j}}},
   $$
   where $\alpha\in(0,1]$ is a learning rate.
  \end{enumerate}
\end{enumerate}

\uone{After the given number of iterations is finished, we can optionally apply gradient-based local optimization to each element in the set $\set{\bm{\theta}\unum{t,i,j}}_{i,j}$ of the approximated local minima.}

\begin{figure}[!tb]
    \centering
    \includegraphics[width=1\linewidth]{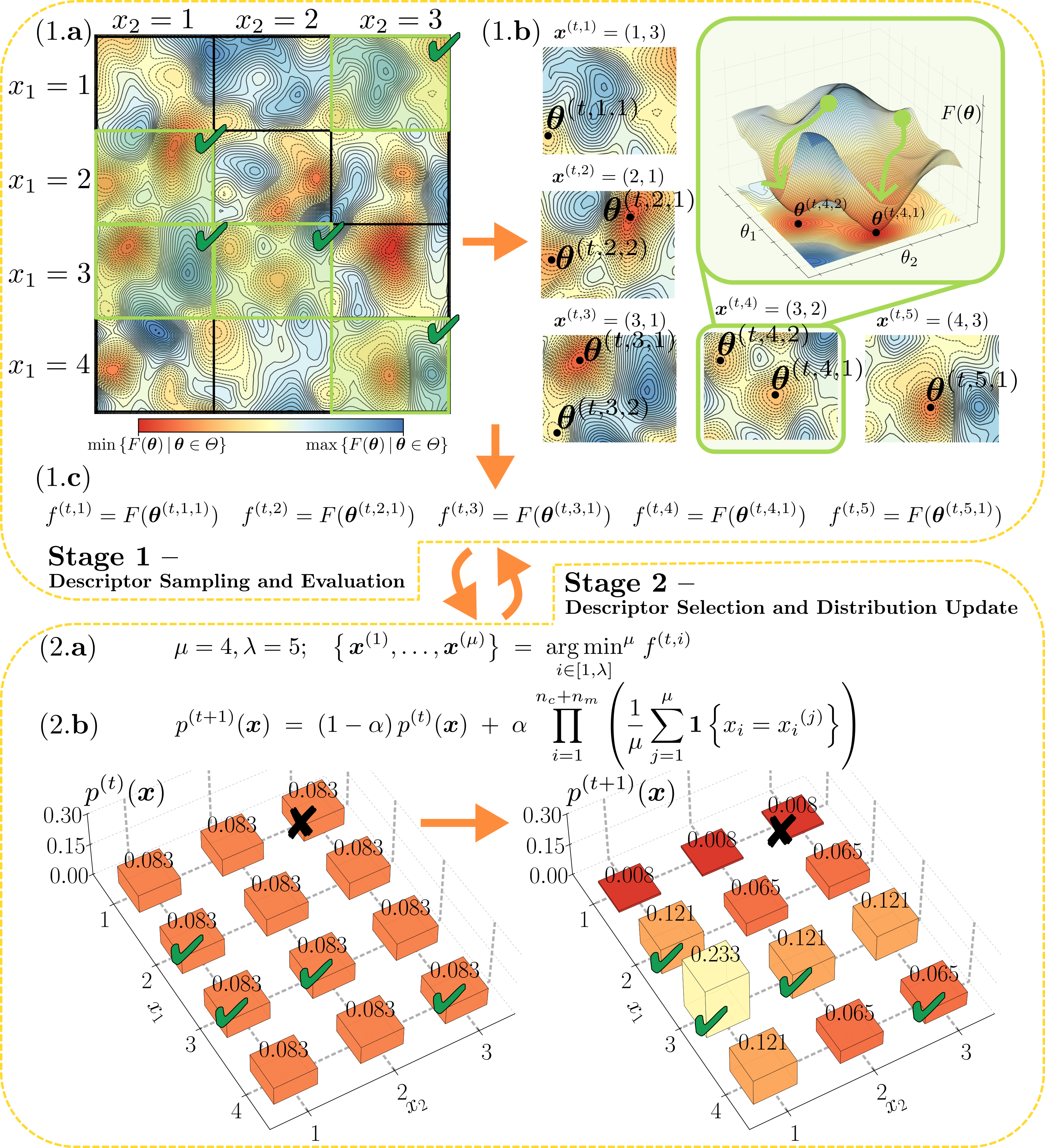}
\caption{
The Lens-Descriptor-Guided Evolutionary Algorithm (LDG-EA) applied to a two-dimensional test function.
Panels (1.\textbf{a}), (1.\textbf{b}), etc. match the corresponding stages described in Sec.~\ref{sec:stages}.
In panel (1.\textbf{a}), the descriptors are indicated by black squares over the domain of the test function.
Green ticks in Stage 1 denote sampled descriptors, while in Stage 2 green ticks denote descriptors selected after evaluation of the descriptor-level objectives $f^{(t,i)}$; a black cross marks a descriptor that is not selected.
The two plots at the bottom visualize the update of the probabilistic distribution over descriptors, $p\unum{t}(\bm{x})$, according to the formula displayed in panel (2.\textbf{b}).
}
    \label{fig:ldg-ea}
\end{figure}

\subsection{Learning Descriptor Models}
After each update of $p\unum{t+1}$, we refine the internal models that generate descriptor components (signs of curvature and material indices, \uone{the thicknesses are ommited due to the resons discussed in Sec.~\ref{sec:formalization}}) by fitting probability distributions conditioned on the selected descriptors. 
Concretely, if a descriptor $\bm{x}$ decomposes as
$$
\bm{x} = (\sgn(\kappa_1),\dots,\sgn(\kappa_{n_c}),\, m_1,\dots,m_{n_m}),
$$
we update:
\begin{itemize}
 \item A Bernoulli model for each curvature sign $\sgn(\kappa_j)$ based on its empirical frequency in the selected set of $\mu$ best performing descriptors.
 \item A categorical model for each material index $m_j$ based on its empirical frequency in the selected set of $\mu$ best-performing descriptors.
\end{itemize}
These learned models are then used to sample better descriptors in subsequent iterations, therefore, biasing the search toward promising structural patterns.
\uone{However, the total number of descriptors is too large to sample them all, so the best learned descriptors are just an approximation of the true best descriptors.}

\subsection{Internal optimization algorithm}
\label{sec:internal-optimizer}

We employ the Hill-Valley Evolutionary Algorithm (HV-EA)~\cite{maree2018real} to discover a diverse set of high-quality solutions sharing the same descriptor $\bm{x}$. 
This algorithm is used on Stage 1 (b) in the proposed method LDG-EA (see Sec.~\ref{sec:stages}).
By construction, each HV-EA run holds the material indices fixed and varies only the surface curvatures $\bm{\kappa}$ and thicknesses $\bm{d}$ within the sign‐constraints implied by $\bm{x}$.
Specifically, for each curvature component $\kappa_i$:
\[
 \begin{cases}
  \kappa_i \le -4\ \text{mm}, & \text{if }\sgn(x_i) = -1,\\
  \kappa_i \ge +4\ \text{mm}, & \text{if }\sgn(x_i) = +1,
 \end{cases}
\]
where the threshold of $4\,$mm reflects our manufacturing limits~\cite{antonov2023new,antonov2024quality}.

\paragraph{HV-EA Overview.}
HV-EA partitions the descriptor‐constrained search space into niches via hill-valley clustering, then runs an \uone{another optimization method} within each niche to approximate local minima \uone{over a given budget}. 
This two‐phase strategy balances exploration, namely identifying distinct attraction basins within the considered descriptor, with exploitation, namely refining within each basin.

\paragraph{Internal Optimizer of HV-EA: CMSA-ES.}
We adopt the Covariance Matrix Self‐Adaptation Evolution Strategy (CMSA‐ES)~\cite{beyer2008covariance} as the internal single‐mode optimizer within HV‐EA, instead of the AMaLGaM‐Univariate algorithm~\cite{bosman2013benchmarking} suggested by the HV‐EA authors~\cite{maree2018real,maree2021two}. 
In our preliminary experiments, CMSA‐ES generated candidate solutions that lie slightly further from the exact local minima than those of AMaLGaM.
Yet solutions by CMSA‐ES remain sufficiently close to reveal the structure of the lens and enable rapid convergence of subsequent gradient‐based optimization. 
By contrast, AMaLGaM’s finer‐scale focus on approximating Gaussian‐mutation iso‐contours can place solutions much closer to the minimum, but this extra precision is redundant once precise gradients are available and only serves to consume additional computational budget.
Moreover, the lightweight self‐adaptation mechanism of CMSA‐ES yields faster initial progress on the highly rugged landscapes of optical design problems.

\paragraph{Archive Maintenance.}
The algorithm enforces a \emph{quality window} of width $w$ on each niche archive $\bm{A}^{(t,i)}$. Concretely, let
\[
 \bm{A}^{(t,i)} = \bigl\{\bm{\theta}^{(t,i,1)},\dots,\bm{\theta}^{(t,i,n)}\bigr\},
 \quad
 F\bigl(\bm{\theta}^{(t,i,1)}\bigr)
 \;\le\;
 \cdots
 \;\le\;
 F\bigl(\bm{\theta}^{(t,i,n)}\bigr),
\]
and define
\[
 F_{\min}
 \;=\;
 F\bigl(\bm{\theta}^{(t,i,1)}\bigr).
\]
We require for all $k$ that
\[
 F\bigl(\bm{\theta}^{(t,i,k)}\bigr)
 \;\le\;
 F_{\min} + w.
\]
When a new candidate $\bm{\theta}'$ with value $F(\bm{\theta}')$ is generated, we insert it into the sorted archive and then remove every solution whose objective exceeds $F_{\min}+w$. This mechanism ensures that the objective value span in each archive 
$\bm{A}^{(t,i)}$ remains bounded by $w$, effectively filtering out lower‐quality local minima. The window width $w$ is calibrated from known reference designs: it is large enough to allow a diverse set of near‐optimal solutions, yet small enough to exclude any a priori inferior configurations.

\paragraph{Stopping Criterion of CMSA‐ES.}
Each CMSA‐ES instance terminates as soon as \emph{any} one of the following conditions is met:
\begin{enumerate}
 \item \textbf{Evaluation budget:} The total number of objective evaluations reaches $T_{\mathrm{HV\text{-}EA}}.$
 \item \textbf{Parameter‐change tolerance:} The change in the mean search vector between two successive generations of CMSA-ES falls below $\varepsilon_{\text{param}}:$
 \[
  \bigl\lVert \bm{\mu}^{(g+1)} - \bm{\mu}^{(g)} \bigr\rVert_2 \le \varepsilon_{\text{param}}.
 \]
 \item \textbf{Function‐value tolerance:} The absolute improvement in the best objective value from one generation to the next is less than $\varepsilon_{\text{fun}}:$
 \[
  \bigl|F_{\min}^{(g+1)} - F_{\min}^{(g)}\bigr| \le \varepsilon_{\text{fun}},
 \]
 where $F_{\min}^{(g)}$ is the best fitness in generation $g$.
 \item \textbf{Fitness‐history tolerance:} Over a sliding window of the last $L_{\text{CMSA-ES}}$ generations, no improvement exceeds 
 $\varepsilon_{\text{hist}}$. Equivalently, the range or standard deviation of 
 $$\{F_{\min}^{(g-L+1)},\dots,F_{\min}^{(g)}\}$$ falls below $\varepsilon_{\text{hist}}$.
\end{enumerate}
In our implementation, we use the default HV‐EA values for 
$\varepsilon_{\text{param}}$, $\varepsilon_{\text{fun}}$, $\varepsilon_{\text{hist}}$
and fix the same total budget limit for HV $T_{\text{HV-EA}}$ to enforce a uniform computational effort across all descriptor‐constrained searches.
The specific values of these and other constants used in the implementation of the algorithms are provided in Table~\ref{tab:constants}.

\uone{In our experiments, the evaluation budget is exhausted only once per HV-EA run. In almost all the runs, the stopping criterion per niche is the parameter-change tolerance.}

\subsection{Stopping Criterion of LDG-EA}
The \uone{proposed algorithm (external optimizer)} terminates when one of the following conditions is met:
\begin{enumerate}
 \item A maximum number of iterations $I_{\text{LDG-EA}}$ is reached.
 \item The improvement in the best objective value over the last $L_{\text{LDG-EA}}$ iterations falls below a tolerance $\varepsilon$.
 \item The distribution $p\unum{t}(\bm{x})$ has nearly converged, where the convergence is measured according to the approximated Kullback-Leibler (KL) divergence~\cite{shlens2014notes} to $p\unum{t-1}$.
\end{enumerate}

This two‐stage, distribution‐adaptive approach balances the exploration of the descriptor space with the exploitation of high‐quality design patterns, enabling efficient discovery of diverse, high‐performance optical lens configurations.

\uone{In our experiments, we run until the maximum number of iterations exceeds the given in advance computational budget.}

\section{Numerical validation}

\subsection{Baseline Evolutionary Algorithm}
\label{sec:baseline}

To emphasize the benefits of descriptor‐guided search, we implement a baseline algorithm that jointly changes all design parameters without intermediate descriptor learning for integer variables. Specifically, we employ the Covariance Matrix Adaptation Evolution Strategy (CMA‐ES)~\cite{hansen2001completely} augmented with a restart mechanism~\cite{hansen2009benchmarking} and an integer‐handling strategy for the material variables~\cite{marty2024lb}.

\paragraph{Joint Parameter Optimization.} 
Unlike LDG‐EA, which constrains each sub‐search to a fixed behavior descriptor, the baseline CMA‐ES treats the entire parameter vector
$
\bm{\theta} = \br{\bm{\kappa}, \,\bm{d}, \,\bm{m}}
$
as continuous (for $\bm{\kappa},\bm{d}$) and integer‐valued (for $\bm{m}$). During each generation, real‐valued samples for $\bm{m}$ are rounded to the nearest valid material index, following the procedure in~\cite{marty2024lb} to maintain feasibility, ensure the absence of premature convergence, and avoid disruptive integer perturbations.

\paragraph{Restart Strategy and Stopping Conditions.} 

The baseline CMA‐ES is given a total budget of $T_{\text{CMA-ES}} = \lambda \cdot T_{\text{HV-EA}} \cdot I_{\text{LDG-EA}}$ evaluations of the objective function $F$. The algorithm starts with an identity covariance matrix and runs until it meets one of its built-in stopping conditions, analogical to those for CMSA-ES: 
(i) parameter‐change falls below $\varepsilon_{\text{param}}$, 
(ii) function‐value improvement falls below $\varepsilon_{\text{fun}}$, or 
(iii) a maximum number of iterations is reached. 
Note that we do not consider higher-dimensional variants in this setting, because a change in the integer variables typically induces a much larger jump in the objective value than a small perturbation of the continuous variables. Consequently, near termination we expect only the continuous variables to keep changing, while the integer part remains fixed until a successful stopping condition is met.
If the algorithm converges before spending all $T_{\text{CMA-ES}}$ evaluations, we restart it with the remaining budget and parameters following BIPOP restart scheme~\cite{hansen2009benchmarking}. This restart loop continues until the full $T_{\text{CMA-ES}}$ evaluations are used.

\subsection{Example Lens System}
\label{sec:example-lens}

\begin{figure}
 \centering
 \includegraphics[width=0.8\linewidth]{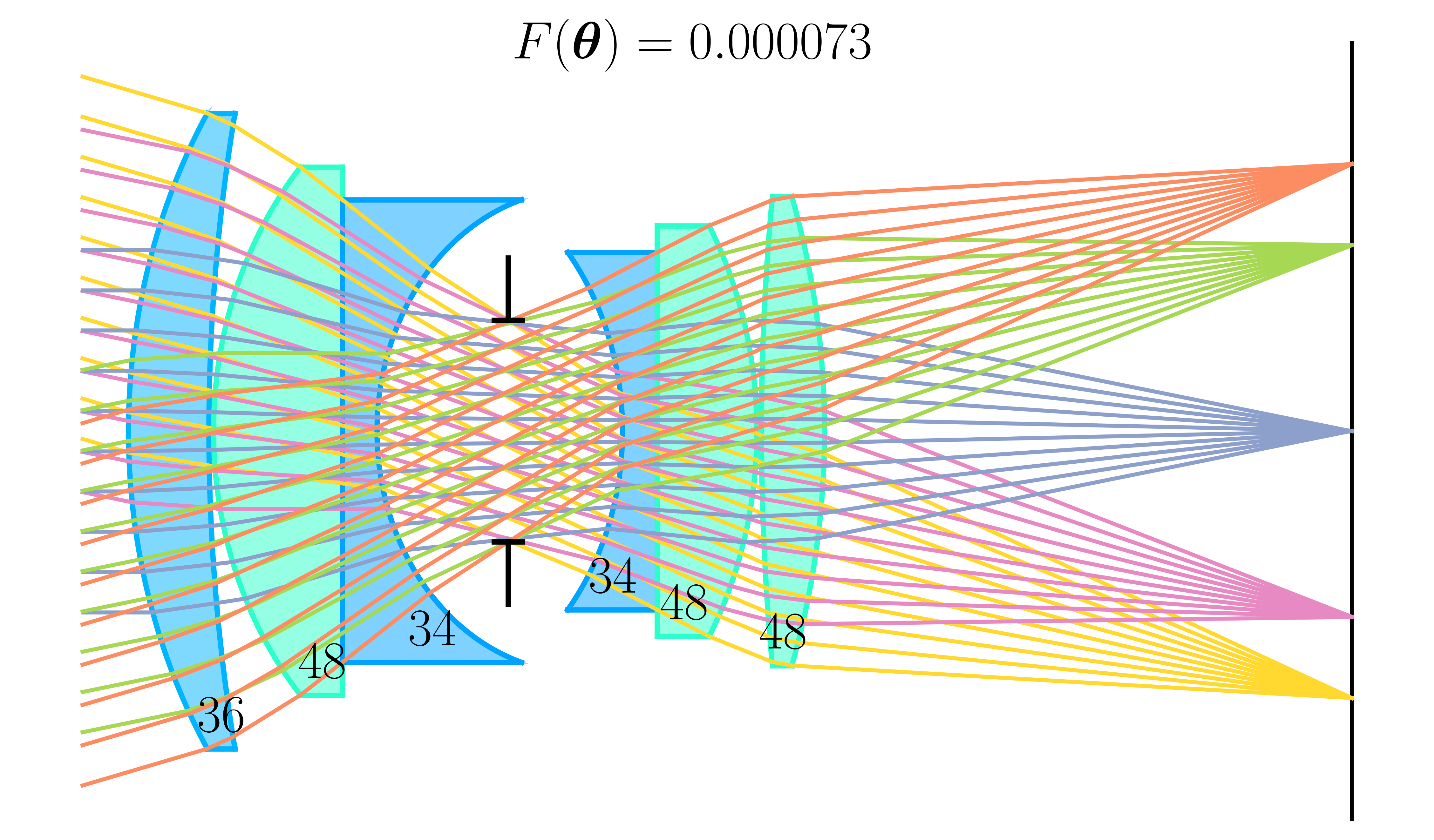}
 \caption{Reference Double‐Gauss lens system. Element glasses are labeled by Schott catalog code and colored by their sorted refractive index at the d‐line (587.6 nm) over the catalog.}
 \label{fig:double-gauss}
\end{figure}

To demonstrate the capabilities of our proposed algorithms, we apply them to a classic Double‐Gauss lens. Using the Double‐Gauss as a \emph{reference template} (Figure~\ref{fig:double-gauss}), we seek to generate alternative configurations that match its principal optical specifications while minimizing RMS spot size objective.

\begin{itemize}
 \item \textbf{Effective focal length:} $f_\text{eff}=95.5\,$mm;
 \item \textbf{Full field of view:} $28^\circ$;
 \item \textbf{Entrance pupil diameter:} $D_\text{EP}=33.33\,$mm;
 \item \textbf{Glass catalog:} Schott~\cite{schott2025} catalog consisting of 120 glasses;
 \item \textbf{Illumination:} polychromatic design over standard spectral lines.
\end{itemize}

In our reference design, the six‐element Double‐Gauss layout features a positive first curvature and successive refracting surfaces in a fixed sequence. We preserve this topology and optimize a broad set of parameters-surface curvatures, element thicknesses, inter-element spacings, and material selections, while keeping the distances from the aperture stop to the neighboring lens vertices constant. The resulting high dimensionality (24 variables in total) and complex interactions between them yield a vast number of local minima \uone{(in the search space, not per descriptor)} with nearly identical RMS‐spot sizes, making this lens system a sufficiently good example for assessing the descriptor‐guided diversity of LDG‐EA. 

During LDG-EA execution we use paraxial solve to find the distance to the sensor from the last surface~\cite{kidger2004intermediate}. However, during the optional subsequent gradient-based optimization, we treat it as a variable.

\paragraph{Double-Gauss Objective Function.} 
We define the overall objective 
\[
F(\bm{\theta})
\;=\;
\underbrace{\bigl(\mathrm{RMS}(\bm{\theta})\bigr)^2}_{\text{imaging quality}}
\;+\;
\sum_{k=1}^{5} w_k\,P^2_k(\bm{\theta}),
\]
where each $P_k(\bm{\theta}) \in \R$ is a penalty term and $w_k>0$ its corresponding weight. We implemented the following penalties:
\begin{itemize}
 \item $P_1$: \emph{vignetting/refracted‐ray penalty}. A large penalty is applied if any ray is vignetted or undergoes total internal reflection during the ray trace. We used $w_1 = 10.$
 \item $P_2$: \emph{negative optical path penalty}. Enforces non‐negative optical path length inside each lens element by penalizing any back‐tracking rays at the element edges. We used $w_2 = 1.$
 \item $P_3$: \emph{minimum‐thickness/air‐gap penalty}. Penalizes element thicknesses or inter‐element air gaps that fall below manufacturable limits. We used $w_3 = 1.$
 \item $P_4$: \emph{free-working distance penalty}. Penalizes the distance from the last surface to the image sensor when it deviates below a specified minimum. We used $w_4 = 1.$
 \item $P_5$: \emph{focal‐length penalty}. Penalizes deviation of the effective focal length from the user‐specified target. We used $w_5 = 1.$
\end{itemize}
The weights $w_k$ are chosen to enforce hard constraints (e.g.\ vignetting) with a relatively large $w_k$, while softer constraints (e.g.\ focal length tolerance) use moderate weights. This composite merit function balances image‐quality optimization against practical design constraints in a single, differentiable objective. We implemented an automatic differentiable optical simulator to compute the value of this function $F$. The implementation follows the standard ray-tracing algorithm for geometric optics~\cite{kidger2001fundamental} and uses automatic differentiation for efficient gradient computations~\cite{wang2022differentiable}.
In our current implementation, we assume the first curvature is always positive. As a result, we disregard descriptors with negative first curvature. We intend to account for negative first curvatures in our future work.

\subsection{Results of Double-Gauss Lens Topology Optimization}

\paragraph{Wall‐clock Time.} 

\uone{All experiments were executed on a Rocky Linux 9 cluster node with 512~GB RAM and two AMD EPYC 7702 CPUs (64 cores each; 128 cores / 256 hardware threads total). We configured the runtime so that each optimization instance within a single behaviour descriptor used at most two CPU threads; consequently, a full LDG-EA run used at most 100 hardware threads concurrently. Peak memory consumption was approximately 12~GB.}

In our LDG‐EA implementation, \uone{running local optimization within a single behavior descriptor} requires approximately 4.5 minutes of wall‐clock time. Since the $\lambda$ descriptor evaluations within each generation are independent, full parallelization would limit the per‐generation runtime to this single‐descriptor time. Over 15 generations, the total runtime was
\[
 15 \times 4.5\ \mathrm{min} \approx 1\ \mathrm{h}.
\]
We allocated the same total evaluation budget to the baseline algorithm. 
Because the baseline lacks a parallel component, this resulted in an approximately 50 times longer wall-clock runtime. 
Due to the resulting runtime of over 50 hours, we executed the baseline only once, whereas we performed five independent runs of the proposed LDG-EA.

\paragraph{Diversity of the Approximated Minima.} 
Across all runs, LDG-EA produced on average \(14741\) local-minimum candidates (standard deviation \(\approx 2049\)) over the full evaluation budget. 
These candidates were distributed over \(636\) distinct behavior descriptors on average (standard deviation \(\approx 37\)), which is below the theoretical maximum of \(\lambda \times I_{\text{LDG-EA}} = 750\). 
This gap indicates that the descriptor distribution stabilized during the search, leading to repeated sampling of a subset of descriptors. 
By Definition~\ref{def:descr-eq}, solutions from different descriptors are never equivalent. 
Moreover, within each descriptor, LDG-EA identifies multiple pairwise-distinct local minima (sharing the same material indices and curvature-sign pattern, but differing in continuous parameters), as enforced by the Hill-Valley predicate in HV-EA.

In contrast, our CMA‐ES baseline located only about $400$ solutions overall, of which just $181$ achieved objective values at least as good as the worst LDG‐EA solution. Although each baseline solution used a unique glass assignment, the joint optimization approach was less systematic and discovered far fewer high‐quality designs under the same computational budget. 

\paragraph{Quality of the Approximated Minima.} 
Across all runs, LDG‐EA produced solutions whose objective values ranged from $7\times10^{-4}$ up to $2.0$. For reference, the unrefined Double‐Gauss template (Figure~\ref{fig:double-gauss}) has $F=5\times10^{-4}$, which improves to $7\times10^{-5}$ after gradient-based optimization of its final surface and sensor distance. After applying our full LDG‐EA pipeline, including gradient‐based optimization, our best candidate achieved $F=3\times10^{-4}$.

\begin{figure*}[!tb]
  \centering
  \includegraphics[width=1\linewidth]{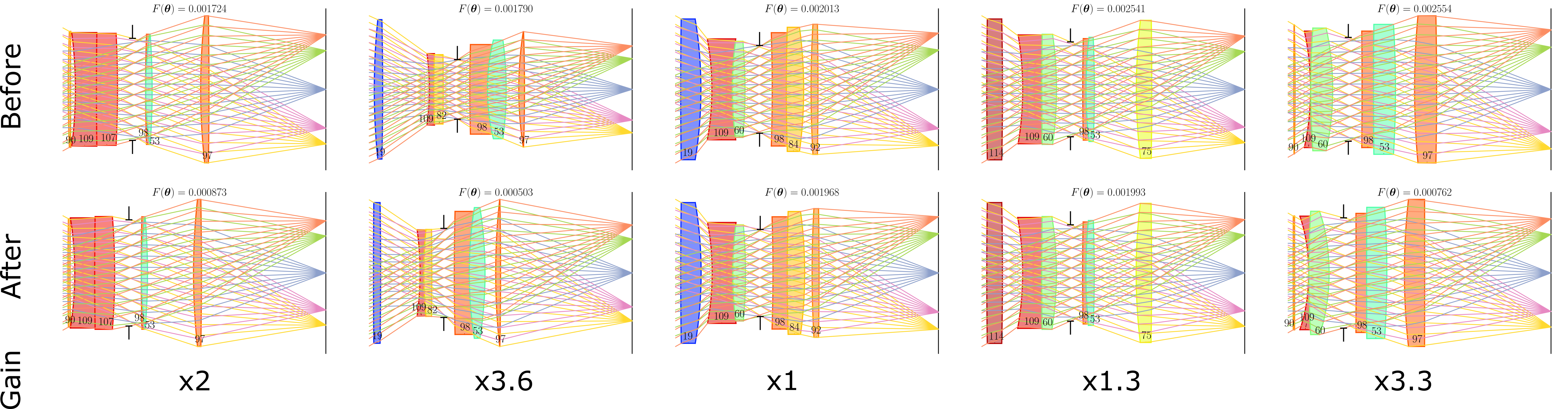}
  \caption{Top five lens designs from one randomly selected LDG-EA run (first row) and their locally refined counterparts obtained with BFGS (second row). The third row reports the additional post-optimization gain in \(F\), computed as \(F(\bm{\theta}^{\text{(before)}})/F(\bm{\theta}^{\text{(after)}})\). Element glasses are labeled by Schott catalog code and colored by their sorted refractive index at the d‐line (587.6 nm) over the catalog.}
  \label{fig:optimized-example}
\end{figure*}

By contrast, the CMA‐ES baseline’s best solution reached only $F=1\times10^{-3}$, and its second‐best was already worse than 36 distinct LDG‐EA solutions (from the same randomly selected run). In total, just 181 of the CMA‐ES outputs matched or exceeded the worst LDG‐EA performance.

Finally, we further refined the top five solutions from one randomly selected LDG-EA run (out of five) using the BFGS quasi-Newton method~\cite{nocedal1999numerical}. 
This post-optimization yielded an additional improvement in \(F\) ranging from \(\times1.0\) (negligible) to \(\times3.6\).
Figure~\ref{fig:optimized-example} compares these five LDG-EA solutions with their locally refined counterparts.
In the BFGS stage, we optimized the distance to the sensor and all surface curvatures, while keeping all remaining design parameters fixed.
Each BFGS run terminated after 1000 iterations or once the gradient norm fell below \(10^{-6}\).
Overall, the parameters of solutions changed only marginally as seen in Figure~\ref{fig:optimized-example}, suggesting that LDG-EA without the gradient-based refinement already produces designs close to locally optimal.

Even so, none of the solutions in all five runs surpassed the tuned reference design’s $7\times10^{-5}$~\uone{(we have not overpassed the baseline)}. However, it does not diminish LDG‐EA’s practical utility: in modern optical‐design workflows, RMS spot size can serve as a heuristic starting point rather than a final target~\cite{devoptical2021rms}, and LDG‐EA is a fast and configurable method to generate high‐quality initial designs. 

\paragraph{Demonstration of Descriptor Learning.}
We demonstrate that LDG-EA learns a non-uniform distribution over the descriptor space that increasingly favors descriptors in which the subsequent HV-EA stage tends to discover higher-quality solutions. 
To isolate this effect, we ablated the first stage of LDG-EA and replaced the proposed descriptor sampling with uniform random sampling. 
We refer to this variant as \emph{Ablated LDG-EA}.

\begin{figure}[!tb]
  \centering
  \begin{tabular}{p{0.95\linewidth}}
       \adjustbox{valign=c, center}{\includegraphics[width=0.6\linewidth]{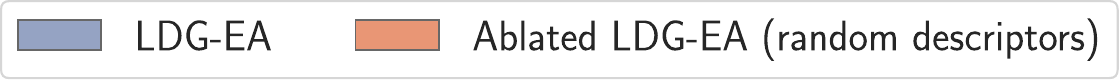}}  \\
       \adjustbox{valign=c, center}{\includegraphics[width=0.9\linewidth]{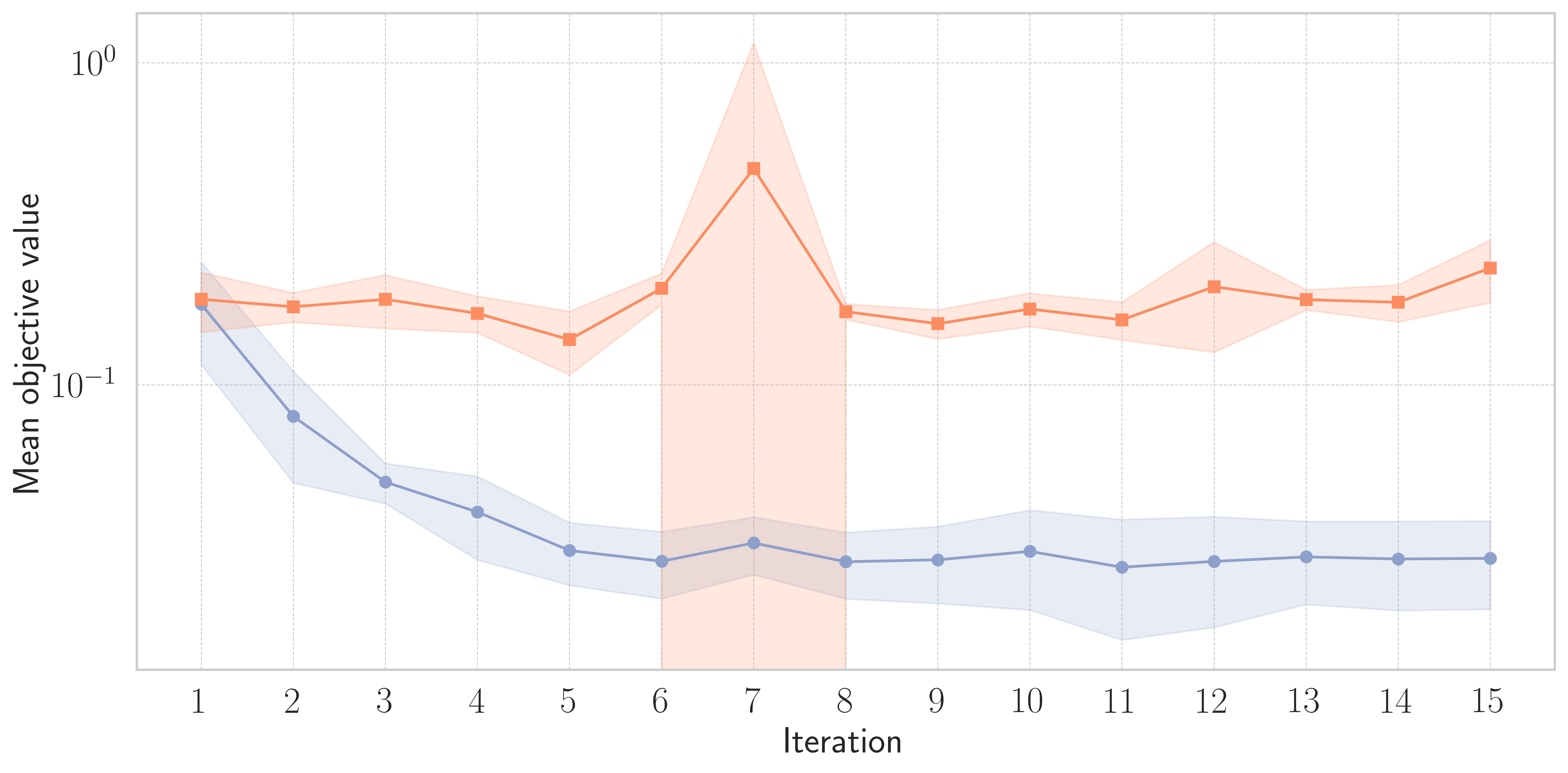}} \\
       \adjustbox{valign=c, center}{(a) Learning curves of the efficient sampling distribution over descriptor space.} \\[2mm]
       \adjustbox{valign=c, center}{\includegraphics[width=0.9\linewidth]{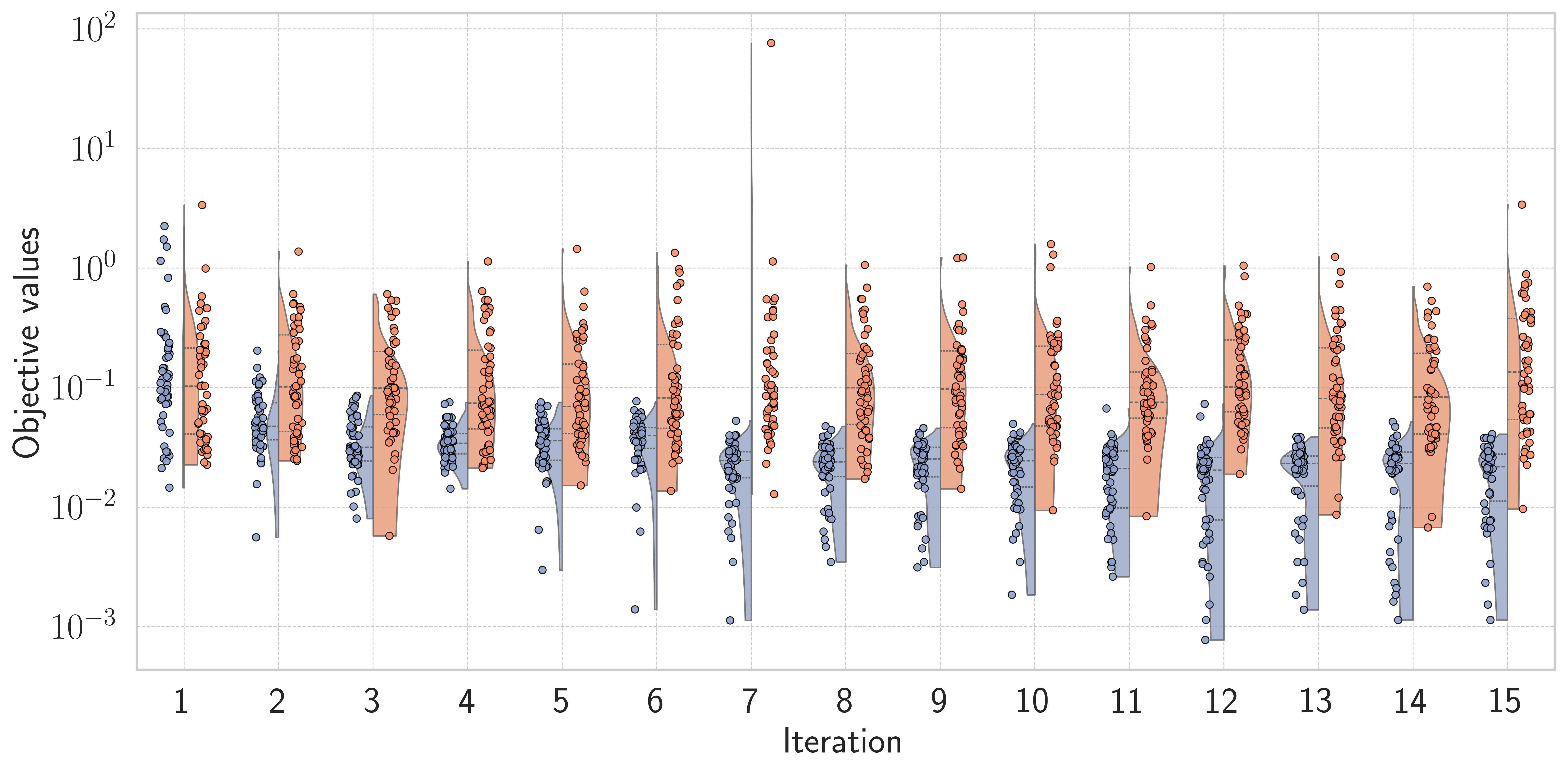}} \\
       (b) Single-run comparison between LDG-EA and Ablated LDG-EA, depicting an Ablated LDG-EA outlier at iteration~7. \\
  \end{tabular}
  \caption{Demonstration that LDG-EA learns a non-uniform distribution over the descriptor space that increasingly favors descriptors in which the subsequent HV-EA stage tends to discover higher-quality solutions. The figure compares the average solution quality produced by LDG-EA and Ablated LDG-EA at each LDG-EA iteration.}
  \label{fig:learning}
\end{figure}

Figure~\ref{fig:learning} compares the average solution quality produced by LDG-EA and Ablated LDG-EA at each LDG-EA iteration (blue: LDG-EA; salmon: Ablated LDG-EA). 
Panel~(a) reports, across five independent runs, the average of the mean in every iteration of LDG-EA (best solutions in $\lambda$ descriptors) in five runs of LDG-EA. The shaded area represents the standard deviation of those means.
Panel~(b) shows a representative single-run comparison. Each dot corresponds to the best solution returned after optimization within a descriptor; we plot all \(\lambda\) dots per iteration for both algorithms.

As the blue mean curve (Figure~\ref{fig:learning} panel a) and the downward shift of the violin plots (Figure~\ref{fig:learning} panel b) illustrate, both the typical and the worst‐case solution quality improve rapidly in the first few generations, confirming that the descriptor distribution quickly adapts to favor high‐yield regions of the design space. After generation ~5, improvement slows, indicating that exploration is giving way to the exploitation of a smaller set of promising descriptors. The occasional outlier boxes below the main trend show that LDG-EA still discovers novel, superior designs even in later generations. Together, these patterns validate that our two‐stage learning mechanism effectively concentrates search effort on descriptors most likely to produce high-quality local minima, while also suggesting that a reduced learning rate or occasional random descriptor injections could help to sustain exploration, overcoming performance plateaus. \uone{The percentage of investigated descriptors is very small, since in total there are at least $10^{12}$ different descriptors; therefore, some even better descriptors can still be found.}

\section{Conlusion}
\label{sec:conclusion}

We have presented the \emph{Lens Descriptor-Guided Evolutionary Algorithm} (LDG-EA), a two‐stage, descriptor‐driven framework for discovering diverse, high-quality local optima in complex optical‐design spaces. By iteratively sampling behavior descriptors, exploiting them via HV-EA powered by CMSA-ES, and then updating the descriptor distribution, LDG-EA dynamically concentrates search effort on the most promising regions of the design space.

Applied to the six‐element Double‐Gauss topology, LDG-EA demonstrated:
\begin{itemize}
 \item \textbf{High diversity:} $\sim14741$ local‐minimum candidates grouped into $636$ distinct descriptors (in average over five runs), compared with only $\sim400$ solutions from a joint CMA-ES baseline.
 \item \textbf{Competitive quality:} \uone{LDG-EA produced} best post‐processed with gradient optimization RMS spot size $F=3\times10^{-4}$, outperforming CMA-ES’s best $F=1\times10^{-3}$ and approaching the tuned reference’s $7\times10^{-5}$.
 \item \textbf{Efficient learning:} rapid descent \uone{of LDG-EA} in mean and worst‐case $F$ across the first five generations, with the continued discovery of outlier improvements in later iterations.
\end{itemize}

LDG-EA’s inherent parallelism makes it applicable on modern computing infrastructures. Furthermore, by treating RMS spot size as a heuristic initialization metric rather than a terminal criterion~\cite{devoptical2021rms}, LDG‐EA provides a flexible starting point for subsequent gradient‐based optimization or for full end-to-end diffraction‐limited design. 

\paragraph{Future Work.} 
In future work, we plan to investigate adaptive learning-rates $\alpha$ and periodic random descriptor injections to maintain exploration in later generations. 
We also plan to extend LDG-EA to additional lens topologies, incorporate manufacturing-tolerance models, and support multi-objective optimization criteria such as total system cost. 
Leveraging existing designs reported in patents as priors or warm-starts may further improve design quality and diversity, which we leave for future study.
Finally, applying LADE’s~\cite{lin2025landscape} and NetCDE~\cite{chen2023network} mechanisms to generate a larger set of candidate peaks could improve the chances of finding superior designs. We leave the integration of those algorithms into our LDG-EA framework for future work. 

\section*{Declaration of generative AI and AI-assisted technologies in the manuscript preparation process}

During the preparation of this work, the authors utilized ChatGPT to correct grammatical errors and rephrase some sentences to enhance clarity. After using this service, the authors reviewed and edited the content as needed and take full responsibility for the content of the published article.

\section*{Acknowledgments}
This work was supported by High Tech Holland, project number TKI HTSM/22.0029.

\bibliographystyle{ieeetr}
\bibliography{myreferences}

\newpage
\appendix

\section{Constants}

\begin{table}[htbp]
 \centering
 \caption{Key constants used in the implementation of the algorithms}
 \label{tab:constants}
 \begin{tabular}{@{}llp{0.6\linewidth}@{}}
  \toprule
  Symbol & Value & Description \\ 
  \midrule
  $\lambda$          & 50      & Number of descriptors sampled per iteration \\ 
  $\mu$            & 5       & Number of top descriptors selected for update \\ 
  $\alpha$          & 1     & Learning rate for descriptor distribution update \\ 
  $w$             & 0.50     & Quality‐window width (allowable $F$‐span in archive) \\ 
  $I_{\text{LDG-EA}}$     & 15      & Number of iterations in the algorithm LDG-EA \\
  $B = T_{\text{HV-EA}}$ & $10^5$    & Total budget of $F$ evaluations for each HV‐EA run \\ 
  $T_{\text{CMA-ES}}$ & $50 \cdot 15 \cdot 10^5$ & Total budget of $F$ evaluations for each CMA-ES run \\
  $\varepsilon_{\mathrm{param}}$ & $10^{-10}$ & CMSA‐ES and CMA-ES mean‐vector update tolerance \\ 
  $\varepsilon_{\mathrm{fun}}$  & $10^{-10}$ & CMSA‐ES and CMA-ES objective‐change tolerance \\ 
  $\varepsilon_{\mathrm{hist}}$ & $10^{-5}$ & CMSA‐ES and CMA-ES fitness‐history stagnation tolerance \\ 
  $\kappa_{\min}$       & $4\,$mm    & Absolute curvature threshold (manufacturing limit) \\ 
  \bottomrule
 \end{tabular}
\end{table}

\end{document}